# Self-supervised learning of hologram reconstruction using physics consistency

Luzhe Huang[1,2,3,*], Hanlong Chen[1,2,3,*], Tairan Liu[1,2,3], Aydogan Ozcan[1,2,3,4]

1 Electrical and Computer Engineering Department, University of California, Los Angeles, CA 90095, USA

2 Bioengineering Department, University of California, Los Angeles, CA 90095, USA

3 California NanoSystems Institute (CNSI), University of California, Los Angeles, CA 90095, USA.

4 David Geffen School of Medicine, University of California Los Angeles, Los Angeles, CA 90095, USA

Corresponding author: ozcan@ucla.edu

* These authors contributed equally to this work

**Abstract**

The past decade has witnessed transformative applications of deep learning in various computational imaging, sensing and microscopy tasks. Due to the supervised learning schemes employed, these methods mostly depend on large-scale, diverse, and labeled training data. The acquisition and preparation of such training image datasets are often laborious and costly, also leading to biased estimation and limited generalization to new sample types. Here, we report a self-supervised learning model, termed GedankenNet, that eliminates the need for labeled or experimental training data, and demonstrate its effectiveness and superior generalization on hologram reconstruction tasks. Without prior knowledge about the sample types to be imaged, the self-supervised learning model was trained using a physics-consistency loss and artificial random images that are synthetically generated without any experiments or resemblance to real-world samples. After its self-supervised training, GedankenNet successfully generalized to experimental holograms of various unseen biological samples, reconstructing the phase and amplitude images of different types of objects using experimentally acquired test holograms. Without access to experimental data or knowledge of real samples of interest or their spatial features, GedankenNet's self-supervised learning achieved complex-valued image reconstructions that are consistent with the Maxwell's equations, and its output inference and object solutions accurately represent the wave propagation in free-space. GedankenNet framework also exhibits resilience to random, unknown perturbations in the physical forward model, including changes in the hologram distances, pixel size and illumination wavelength. This self-supervised learning of image reconstruction tasks creates new opportunities for various inverse problems in holography, microscopy and computational imaging fields.

**Introduction**

Recent advances in deep learning have revolutionized computational imaging, microscopy and holography-related fields, with applications in biomedical imaging[1], sensing[2], diagnostics[3] and 3D displays[4], also achieving benchmark results in various image translation and enhancement tasks, e.g., super-resolution[5–12], image denoising[13–16] and virtual staining[17–23], among others. The flexibility of deep learning models has also facilitated their widespread use in different imaging modalities, including brightfield[24,25] and fluorescence microscopy[8,11,12,15,26,27]. As another important example, digital holographic microscopy (DHM), a label-free imaging technique widely used in biomedical and physical sciences and engineering[28–37], has also significantly benefitted from deep learning and neural networks[4,38–48]. Convolutional neural networks (CNNs)[38–41,43,45,46,49,50] and recurrent neural networks (RNNs)[47,51] have been used for holographic image reconstruction, presenting unique advantages over classical phase retrieval algorithms, such as using fewer measurements and achieving an extended depth-of-field. Researchers also explored deep learning-enabled image analysis[52–57] and transformations[18,19,25,58,59] on holographic images to further leverage the quantitative phase information (QPI) provided by DHM.

In these existing approaches, supervised learning models were utilized, demanding large-scale, high-quality and diverse training datasets (from various sources and types of objects) with annotations and/or ground truth experimental images. For microscopic imaging and holography, in general, such labeled training data can be acquired through classical algorithms that are treated as the ground truth image reconstruction method[38,39,43,47,48,51], or through registered image pairs (input vs. ground truth) acquired by different imaging modalities[8,17,18,25]. These supervised learning methods require significant labor, time, and cost to acquire, align and pre-process the training images, and potentially introduce inference bias, resulting in limited generalization to new types of objects never seen during the training. Generally speaking, existing supervised learning models demonstrated on microscopic imaging and holography tasks are highly dependent on the training image datasets acquired through experiments, which show variations due to the optical hardware, types of specimens and imaging (sample preparation) protocols. Though there were efforts utilizing unsupervised learning[60–66] and self-supervised learning[16,67–69] to alleviate the reliance on large-scale experimental training data, the need for experimental measurements or sample labels with the same or similar features as the testing samples of interest is not entirely eliminated. Using labeled simulated data for network training is another possible solution; however, generating simulated data distributions to accurately represent the experimental sample distributions can be complicated and requires prior knowledge of the sample features and/or some initial measurements with the imaging set-up of interest[6,10,70–73]. For example, supervised learning-based deep neural networks for hologram reconstruction tasks demonstrated decent *internal* generalization to new samples of the same type as in the training dataset, while their *external* generalization to different sample types or imaging hardware was limited[38,46,51].

A common practice to enhance the imaging performance of a supervised model is to apply transfer learning[51,68,74–76], which trains the learned model on a subset of the *new* test data. However, the features learned through supervised transfer learning using a limited training data distribution, e.g., specific types of samples, do not necessarily advance external generalization to other types of samples, considering that the sample features and imaging set-up may differ significantly in the blind testing phase. Furthermore, transfer learning requires additional labor and time to collect fresh data from the new testing data distribution and fine-tune the pre-trained model, which might bring practical challenges in different applications.

In addition to these, deep learning-based solutions for inverse problems in computational imaging generally lack the incorporation of explicit physical models in the training phase; this, in turn, limits the compatibility of the network's inference with the physical laws that govern the light–matter interactions and wave propagation. Recent studies demonstrated physics-informed neural networks (PINNs)[69,77–82], where a physical loss was formulated to train the network in an unsupervised manner to solve partial differential equations. However, PINN-based methods that can match (or come close to) the performance of supervised learning methods have not been reported yet for solving inverse problems in computational imaging with successful generalization to new types of samples.

Here, we demonstrate the first self-supervised learning (SSL)-based deep neural network for hologram reconstruction, which is trained without any experimental data or prior knowledge of the types or spatial features of the samples. We term it GedankenNet as the self-supervised training of our network model is based on randomly-generated artificial images with no connection or resemblance to real samples at the micro- or macro-scale, and therefore the spatial frequencies and the features of these images do not represent any real-world samples and are not related to any experimental set-up. As illustrated in Fig. 1(a), the self-supervised learning scheme of GedankenNet adapts a *physics-consistency* loss between the input synthetic holograms of random, artificial objects, and the numerically predicted holograms calculated using the GedankenNet output complex fields, without any reference to or use of the ground truth object fields during the learning process. After its training, the self-supervised GedankenNet directly generalizes to experimental holograms of various types of samples even though it never saw any experimental data or used any information regarding the real samples. When blindly tested on experimental holograms of human tissue sections (lung, prostate and salivary gland tissue) and Pap smears, GedankenNet achieved better image reconstruction accuracy compared to supervised learning models using the same training datasets. We further demonstrated that GedankenNet can be widely applied to other training datasets, including simulations and experimental datasets, and achieves superior generalization to unseen data distributions over supervised learning-based models.

Since GedankenNet's self-supervised learning is based on a physics-consistency loss, its inference and the resulting output complex fields are compatible with the Maxwell's equations and accurately reflect the physical wave propagation phenomenon in free-space. By testing GedankenNet with experimental input holograms captured at shifted (unknown) axial positions, we showed that GedankenNet does not hallucinate and the object field at the sample plane can be accurately retrieved through wave propagation of the GedankenNet output field, without the need for retraining or fine-tuning its parameters. These results indicate that in addition to generalizing to experimental holograms of unseen sample types without seeing any experimental data or real object features, GedankenNet also implicitly acquired the physical information of wave propagation in free-space and gained robustness towards defocused holograms or changes in the pixel size through the same self-supervised learning process. Furthermore, for phase-only objects (such as thin label-free samples), GedankenNet framework also exhibits resilience to random unknown perturbations in the imaging system, including arbitrary shifts of the sample-to-sensor distances and unknown changes in the illumination wavelength, all of which make its generalization even broader without the need for any experimental data or ground truth labels.

The success of GedankenNet eliminates three major challenges in existing deep learning-based holographic imaging approaches: (1) the need for large-scale, diverse and labeled training data, (2) the limited generalization to unseen sample types or shifted input data distributions, and (3) the lack of an interpretable connection and compatibility between the physical laws/models and the trained deep neural network. This work introduces a promising and powerful alternative to a wide variety of supervised learning-based methods that are currently applied in various microscopy, holography and computational imaging tasks.

**Results**

The hologram reconstruction task, in general, can be formulated as an inverse problem[42]:

$$\hat{o} = \arg\min_{o} L(H(o), i) + R(o)$$

where $i \in \mathbb{R}^{MN^2}$ represents the vectorized $M$ measured holograms, each of which is of dimension $N \times N$ and $o \in \mathbb{C}^{N^2}$ is the vectorized object complex field. $H(\cdot)$ is the forward imaging model, $L(\cdot)$ is the loss function and $R(\cdot)$ is the regularization term. Under spatially and temporally coherent illumination of a thin sample, $H(\cdot)$ can be simplified as:

$$H(o) = f(Ho) + \epsilon$$

where $H \in \mathbb{C}^{MN^2 \times N^2}$ is the free-space transformation matrix[44,83], $\epsilon \in \mathbb{R}^{MN^2}$ represents random detection noise and $f(\cdot)$ refers to the (opto-electronic) sensor-array sampling function, which records the intensity of the optical field.

Different schemes for solving holographic imaging inverse problems are summarized in Fig. 1. Existing methods for generalizable hologram reconstruction can be mainly classified into two categories, as shown in Fig. 1(a): (1) iterative phase retrieval algorithms based on the physical forward model and iterative error-reduction; (2) supervised deep learning-based inference methods that learn from training image pairs of input holograms $i$ and the ground truth object fields $o$. Similar to the iterative phase recovery algorithms listed under (1), deep neural networks were also used to provide iterative approximations to the object field from a batch of hologram(s); however, these network models were iteratively optimized for each hologram batch separately, and cannot generalize to reconstruct holograms of other objects once they are optimized[69,78,80] (see Supplementary Note 2 and Extended Data Fig. 3).

Different from existing learning-based approaches, instead of directly comparing the output complex fields ($\hat{o}$) and the ground truth object complex fields ($o$), GedankenNet infers the predicted holograms $\hat{\imath}$ from its output complex fields $\hat{o}$ using a deterministic physical forward model, and directly compares $\hat{\imath}$ with $i$. Without the need to know the ground truth object fields $o$, this forward model – network cycle establishes a physics-consistency loss ($L_{physics-consistency}$) for gradient back-propagation and network parameter updates, which is defined as:

$$L_{physics-consistency}(\hat{\imath}, i) = \alpha L_{FDMAE}(\hat{\imath}, i) + \beta L_{MSE}(\hat{\imath}, i),$$

where $L_{FDMAE}$ and $L_{MSE}$ are the Fourier domain mean absolute error (FDMAE) and the mean square error (MSE), respectively, calculated between the input holograms $i$ and the predicted holograms $\hat{\imath}$. $\alpha, \beta$ refer to the corresponding weights of each term (see the Methods section for the training and implementation details). The network architecture of GedankenNet is also detailed in the Methods section and Extended Data Fig. 1.

As emphasized in Fig. 1, GedankenNet eliminates the need for experimental, labeled training data and thus presents unique advantages over existing methods. The training dataset of GedankenNet only consists of artificial holograms generated from random images (with no connection or resemblance to real-world samples), which serve as the amplitude and phase channels of the object field (see the Methods section and Fig. 1(b)). After its self-supervised training using artificial images without any experimental data or real-world specimens, GedankenNet can be directly used to reconstruct experimental holograms of various microscopic specimens, including e.g., densely connected tissue samples and Pap smears. This is vastly in contrast to existing supervised learning methods that exhibit limited external generalization to unseen data distributions and new sample types. Furthermore, compared with classical iterative phase retrieval algorithms, GedankenNet (after its one-time training is complete) provides significantly faster reconstructions in a single forward inference without the need for numerical iterations, transfer learning or fine-tuning of its parameters on new testing samples.

To demonstrate these unique features of GedankenNet, we trained a series of self-supervised network models that take multiple input holograms ($M$ ranging from 2 to 7), following the training process introduced in Fig. 1. Each GedankenNet model for a different $M$ value was trained using artificial holograms generated from random synthetic images based on $M$ different planes with designated sample-to-sensor distances $z_i, i = 1,2,\cdots,M$. In the blind testing phase illustrated in Fig. 2(a), $M$ experimental holograms of human lung tissue sections were captured by a lensfree in-line holographic microscope (see Extended Data Fig. 1(b) and the Methods section for experimental details). We tested all the self-supervised GedankenNet models on 94 non-overlapping fields-of-view (FOVs) of tissue sections and quantified the image reconstruction quality in terms of the amplitude and phase structural similarity index measure (SSIM) values with respect to the ground truth object fields (see Fig. 2(b)). The ground truth fields were retrieved by the multi-height phase retrieval (MHPR[84–86]) algorithm using $M = 8$ raw holograms of each FOV. Our results indicate that all the GedankenNet models were able to reconstruct the sample fields with high fidelity even though they were trained using random, artificial images without any experimental data (Fig. 2(c)). Additionally, Fig. 2 demonstrates that the reconstruction quality of GedankenNet models increased with increasing number of input holograms $M$, which inherently points to a general trade-off between the image reconstruction quality and system throughput; depending on the level of reconstruction quality desired and the imaging application needs, $M$ can be accordingly selected/optimized. In addition to the number of input holograms, we investigated the relationship between the sample-to-sensor distances and the reconstruction quality of GedankenNet; see the Extended Data Figure 2 and Supplementary Note 1. Due to the reduced signal-to-noise ratio (SNR) of the experimental in-line holograms acquired at large sample-to-sensor (axial) distances, GedankenNet models trained with larger sample-to-sensor distances exhibit a relatively reduced reconstruction quality compared with the GedankenNet models trained with smaller axial distances.

We also compared the generalization performance of self-supervised GedankenNet models against other supervised learning models and iterative phase recovery algorithms using experimental holograms of various types of human tissue sections and Pap smears; see Fig. 3. Though only seeing artificial holograms of random images in the training phase, GedankenNet ($M = 2$) was able to directly generalize to experimental holograms of Pap smears and human lung, salivary gland and prostate tissue sections. For comparison, we trained two supervised learning models using the same artificial image dataset, including the Fourier Imager Network (FIN)[48] and a modified U-Net[87] architecture (see the Methods section). These supervised models were tested on the same experimental holograms to analyze their external generalization performance. Compared to these supervised learning methods, GedankenNet exhibited superior external generalization on all four types of samples (lung, salivary gland and prostate tissue sections and Pap smears), scoring higher enhanced correlation coefficient (ECC) values (see the Methods section). A second comparative analysis was performed against a classical iterative phase recovery method, i.e., MHPR[84–86]: GedankenNet inferred the object fields with less noise and higher image fidelity compared to MHPR ($M$=2) that used the same input holograms (see Fig. 3(a,c)). In addition, we compared GedankenNet image reconstruction results against deep image prior (DIP) based approaches[69,78,80,88], also confirming its superior performance (see Extended Data Fig. 3 and Supplementary Note 2).

The inference time of each of these hologram reconstruction algorithms is summarized in Extended Data Table 1, which indicates that GedankenNet accelerated the image reconstruction process by ~128 times compared to MHPR ($M$=2). These holographic imaging experiments and resulting analyses successfully demonstrate GedankenNet's unparalleled generalization to experimental holograms of unknown, new types of samples without any prior knowledge about the samples or the use of experimental training data or labels.

GedankenNet's strong external generalization is due to its self-supervised learning scheme that employs the physics-consistency loss, which is further validated by the additional comparisons we performed between self-supervised learning and supervised learning schemes; see Extended Data Fig. 4 and Supplementary Note 3. In this new analysis, we compared GedankenNet and the supervised learning model FIN that were trained with the same artificial hologram datasets generated from random synthetic images (Extended Data Fig. 4a) or natural images from COCO dataset (Extended Data Fig. 4b). The blind testing of these models used experimental holograms of Pap smear samples and lung tissue sections. The results of this comparison (summarized in Extended Data Fig. 4) reveal that the self-supervised learning scheme consistently achieved better reconstruction accuracy and enhanced ECC scores over the supervised learning scheme, further highlighting the superior external generalization of GedankenNet to experimental holograms of new types of samples.

In addition to GedankenNet's superior external generalization (from artificial random images to experimental holographic data), this framework can also be applied to other training datasets. To showcase this, we trained three GedankenNet models using (i) the artificial hologram dataset generated from random images, same as before; (ii) a new artificial hologram dataset generated from a natural image dataset (COCO)[89]; (iii) an experimental hologram dataset of human tissue sections (see Methods for dataset preparation). Each one of these training datasets had ~100 K training image pairs with $M = 2$, $z_1 = 300$ µm and $z_2 = 375$ µm. As shown in Fig. 4, these three individually trained GedankenNet models were tested on four testing datasets, including artificial holograms of (1) random synthetic images and (2) natural images as well as experimental holograms of (3) lung tissue sections and (4) Pap smears. Our results reveal that all the self-supervised GedankenNet models showed very good reconstruction quality for both internal and external generalization; see Fig. 4(a-b). When trained using the experimental holograms of lung tissue sections, the supervised hologram reconstruction model FIN (solid red bar) scored higher ECC values ($p$ value of $7.5 \times 10^{-38}$) than the GedankenNet (solid blue bar) on the same testing set of the lung tissue sections. However, when it comes to external generalization, as shown in Fig. 4(b), GedankenNet (the blue shadow bar) achieved superior imaging performance ($p$ value of $8.5 \times 10^{-10}$) compared to FIN (the red shadow bar) on natural images (from COCO dataset). One can also notice the overfitting of the supervised model (FIN) by the large performance gap observed between its internal and external generalization performance shown with the red bars in Fig. 4(b). On the contrary, the self-supervised GedankenNet trained with artificial random images (the blue bars) showed very good generalization performance for both test datasets covering natural macro-scale images as well as micro-scale tissue images.

To further illustrate the relationship between the training dataset composed of artificial random images and the generalization performance of GedankenNet, we compared the standard GedankenNet model reported earlier (Fig. 3) against a new GedankenNet model trained on artificial random complex fields with *correlated* amplitude and phase channels. As summarized in Extended Data Figure 5 and Supplementary Note 4, this new GedankenNet model trained with correlated amplitude and phase images generalized relatively worse on the same external test datasets compared to the original GedankenNet model that did not use any correlation between the amplitude and phase channels. These results further confirm that the large data variations in the phase and amplitude channels of the artificially generated random training images greatly contribute to the superior generalization of GedankenNet models.

Besides its generalization to unseen testing data distributions and experimental holograms, the inference of GedankenNet is also compatible with the wave equation. To demonstrate this, we tested the GedankenNet model (trained with the artificial hologram dataset generated from random synthetic images) on experimental holograms captured at shifted unknown axial positions $z_1' \cong z_1 + \Delta z \ and \ z_2' \cong z_2 + \Delta z$, where $z_1, z_2$ were the training axial positions and $\Delta z$ is the unknown axial shift amount. The same model as in Fig. 3 was used for this analysis and blindly tested on lung tissue sections (i.e., external generalization). Due to the unknown axial defocus distance ($\Delta z$), the direct output fields of GedankenNet do not match well with the ground truth, indicated by the orange curve in Fig. 5(a). However, since GedankenNet was trained with the physics-consistency loss, its output fields are compatible with the wave equation in free-space. Thus, the object fields at the sample plane can be accurately retrieved from the GedankenNet output fields by performing wave propagation by the corresponding axial defocus distance. After propagating the output fields of GedankenNet by $-\Delta z$ using the angular spectrum approach, the propagated fields (blue curve) matched very well with the ground truth fields across a large range of axial defocus values, $\Delta z$. These results are important because (1) they once again demonstrate the success of GedankenNet in generalizing to experimental holograms even though it was only trained by artificial holograms of random synthetic images; and (2) the physics-consistency based self-supervised training of GedankenNet encoded the wave equation into its inference process so that instead of hallucinating and creating non-physical random optical fields when tested with defocused holograms, GedankenNet outputs correct (physically consistent) defocused complex fields. In this sense, GedankenNet not only exhibits superior external generalization (from experiment- and data-free training to experimental holograms), but also very well generalized to work with defocused experimental holograms. To the best of our knowledge, these features were not demonstrated before for any hologram reconstruction neural network in the literature.

Figure 5(b) reports another example of GedankenNet's superior external generalization and its compatibility with the wave equation. The same trained GedankenNet model of Fig. 5(a) was blindly tested on experimental holograms of unstained (*label-free*) human kidney tissue sections, which can be considered phase-only samples. Besides the success of GedankenNet's generalization to experimental data of biological samples, the results shown in Fig. 5(b) demonstrate GedankenNet's external generalization to another physical class of objects (i.e., phase-only samples) that exhibit different physical properties than the synthetic, artificial random complex fields used in the training, which included random phase and amplitude patterns. Stated differently, although GedankenNet's artificially generated random training images did not include any phase-only objects, it successfully reconstructed the experimental holograms of phase-only objects – the first time that they were seen. Furthermore, similar to Fig. 5(a), we observe in Fig. 5(b) that by digitally propagating the GedankenNet outputs for defocused input holograms of the label-free tissue samples (orange curve) by an axial distance of $-\Delta z$, the resulting phase reconstructions (blue curve) showed good fidelity to the ground truth phase images of the same samples.

In addition to the analyses reported in Figure 5, where the input holograms were defocused by an axial distance of $\Delta z$, we report in Extended Data Figure 6 and Supplementary Note 5 the resilience of GedankenNet models to different experimental input holograms with various pixel pitches that are larger compared to the training pixel pitch. In these results, GedankenNet simultaneously implemented hologram reconstruction and pixel super-resolution without any fine-tuning or retraining of its model, which only utilized artificial random training images at a single pixel pitch (see the Extended Data Figure 6). To further explore the robustness of GedankenNet to variations in other physical features of the acquired holograms of interest, in Supplementary Note 6 and Extended Data Figure 7, we studied the impact of the SNR of the input holograms on the image reconstruction quality, and compared GedankenNet's blind inference results against supervised models, further confirming its superiority and robustness to different sources of noise or perturbations.

This robustness of GedankenNet and its external generalization performance reported in earlier analyses can be further improved using some object priors, such as the assumption of phase-only objects. Different from the earlier models, here the GedankenNet framework uses phase-only artificial random complex fields (with unit amplitude) during its training. Apart from this phase-only object assumption, there is *no* prior knowledge about the sample types to be imaged, and therefore the self-supervised learning model was trained using the same physics-consistency loss and artificial phase-only random objects that were synthetically generated without any experiments or resemblance to real-world samples. This new model, which we termed GedankenNet-*Phase*, exhibits enhanced adaptability to random, unknown perturbations in the forward optical model. Following a similar experimental-data-free training protocol as used in earlier GedankenNet models, GedankenNet-*Phase* was trained on $M = 2$ simulated holograms of artificial phase-only objects; however, these holograms were virtually located at independent random axial positions between 275 μm and 400 μm during the training, aiming to achieve both hologram reconstruction and autofocusing using self-supervised learning based on the same physics-consistency loss (see the Methods and Extended Data Figure 8 for details on the training process and the architecture of GedankenNet-*Phase*). After its training with artificial random data generated without any experiments or resemblance to real-world samples, Fig. 6 demonstrates the experimental hologram reconstruction and autofocusing performance of GedankenNet-*Phase* on unstained (label-free) human kidney tissue sections. Figure 6(a) visualizes GedankenNet-*Phase* outputs corresponding to $M = 2$ input holograms independently captured at arbitrary and unknown axial positions within $[300, 400]$μm, and Fig. 6(b) quantitatively evaluates the reconstruction quality of GedankenNet-*Phase* in terms of phase root mean squared error (RMSE) with respect to the ground truth over a test set of 98 unique FOVs. These results reveal that GedankenNet-*Phase* successfully achieved both autofocusing and hologram reconstruction within its training axial distance range. As expected, the reconstruction quality drops for $z_1 = z_2$ since it corresponds to $M = 1$, deviating from the training of GedankenNet-*Phase,* which used $M = 2$ random input holograms.

The concept of GedankenNet-*Phase* can be further expanded to bring additional resilience to its reconstructions and achieve broader external generalization for other types of perturbations in the physical forward model, such as unknown changes or shifts/drifts in the illumination wavelength. To showcase this, we created an additional model, which we termed

GedankenNet-*Phaseλ* and trained it on $M = 2$ simulated holograms of artificial phase-only random fields, which were illuminated and propagated with random illumination wavelengths between 520 nm and 540 nm (more details are provided in Supplementary Note 7). Apart from the phase-only assumption, the training of GedankenNet-*Phaseλ* did not involve any experiments, data resembling real-world samples, or other prior knowledge about the samples. After its training with artificial random data, GedankenNet-*Phaseλ* generalized to experimental data of unstained human kidney tissue sections and achieved stable reconstruction performance on holograms acquired at various illumination wavelengths, ranging from 500 nm to 560 nm, without knowing what the illumination wavelength is (see Extended Data Figure 9 and Supplementary Note 7). These results and analyses demonstrate the superior robustness of the GedankenNet framework to various sources of perturbations in the physical forward model, while also maintaining its broad external generalization performance.

**Discussion**

In this work, we demonstrated GedankenNet, a self-supervised hologram reconstruction neural network that eliminates the dependence on labeled and experimental training data, and achieves better generalization to unseen data distributions than existing methods. Based on its self-supervised learning scheme and the physics-consistency loss function, GedankenNet is able to implicitly learn the physics of wave propagation and perform hologram reconstruction tasks without any experimental data or prior knowledge of the samples. Stated differently, the training of GedankenNet involves *Gedankenexperiments* (thought experiments) without involving any experimental data or any prior knowledge about real-world samples, and after its training, GedankenNet successfully generalizes to experimental holograms and shows superior reconstruction quality for external generalization compared with supervised learning-based network models. We also demonstrated that GedankenNet outputs are compatible with the wave equation, and it does not hallucinate artificial (non-physical) output fields when defocused holograms are provided as input. These results present an additional degree of successful generalization (beyond experiment- and data-free training to experimental holograms) since during the self-supervised training of GedankenNet we always used $\Delta z = 0$.

Compared with the existing supervised learning methods, GedankenNet has several unique advantages. It eliminates the dependence on labeled experimental training data in computational microscopy, which often come from other imaging modalities or classical algorithms and therefore, inevitably introduce biases for external generalization performance of the trained network. The self-supervised learning scheme of GedankenNet also considerably relieves the cost and labor of collecting and preparing large-scale microscopic image datasets. For the inverse problem of hologram reconstruction, the reported physics-consistency loss that we used in self-supervised learning outperforms traditional structural loss functions commonly employed in supervised learning since they often overfit to specific image features that appear in the training dataset, resulting in generalization errors, especially for new types/classes of samples never seen before (see Extended Data Fig. 4). In general, the residual errors that stochastically occur during the network training would be non-physical errors that are incompatible with the wave equation, e.g., noise-like errors that do not follow wave propagation. In contrast to traditional structural loss functions that penalize these types of residual errors based on the statistics of the sample type of interest (which requires experimental data and/or knowledge about the samples and their features), the physics-consistency loss function that we used focuses on physical inconsistencies, which is at the heart of the superior external generalization of GedankenNet framework since such physical errors are universally applicable, regardless of the type of sample or its physical properties or features. Furthermore, this physics consistency loss benefits from multiple hologram planes (i.e., $M \geq 2$) so that it can also filter out twin-image-related artifacts that would normally appear in conventional in-line hologram reconstruction methods due to lack of direct phase information; stated differently, an artificial twin image that would be superimposed onto the complex-valued true image of the sample would be attacked by our self-consistency loss since it will create physical inconsistencies on at least $M$-1 hologram planes as a result of the wave propagation step for $M \geq 2$ planes. In addition to this, the large degrees of freedom provided by the artificially synthesized image datasets, with random phase and amplitude channels, also contribute to the effectiveness of the GedankenNet framework, as also highlighted in the Results section, Extended Data Figure 5 and Supplementary Note 4. Limited by the optical system, the experimental holographic imaging process applies a low-pass filter to the ground truth object fields. Furthermore, the recurrent spatial features within the same type of samples further reduce the diversity of the experimental datasets. Thus, adapting simulated holograms of random, artificial image datasets presents a more effective solution when access to large amounts of experimental data is impractical (see Fig. 4 and Extended Data Figure 5). In addition, GedankenNet exhibits superior generalization to unseen data distributions than supervised models, and achieves better holographic image reconstruction for unseen, new types of samples (see e.g., Figs. 3-4).

During the training phase of GedankenNet, the physical forward model is given to the network as part of the *Gedankenexperiments*. However, perturbations, i.e., the mismatch between the *a priori* forward model and the *a posteriori* model in the experiments, could impact the performance of the learned GedankenNet. These sources of perturbations often include: (1) the measurement noise $\epsilon$, (2) the modeling error of the sampling function $f$ and

(3) the error of the transfer function $H$. The first two sources can come from a combination of factors, e.g., thermal and shot noise, sensor nonlinearity, aberrations, etc., and can be properly handled by setting regularization terms in the loss function, e.g., the total variation (TV) loss[90]. The last source of perturbations may result from the assumptions when establishing the forward model and errors in the key parameters of the holographic imaging system, e.g., the sample-to-sensor distances, the illumination wavelength, the pixel pitch, etc. Through the self-supervised learning process, GedankenNet intrinsically acquired robustness to various types of random perturbations in the physical forward model (see e.g., Fig. 5 and Extended Data Fig. 6) and implicitly learned the physics of wave propagation. Furthermore, we demonstrated that the GedankenNet framework could adapt to more complicated random, unknown perturbations, as shown with GedankenNet-*Phase* and GedankenNet-*Phaseλ*, which used artificial phase-only random training images that were synthetically generated without any experiments or resemblance to real-world samples. These observations align with earlier reports, which showed self-supervised models to be more robust to adversarial attacks and input image/data corruptions and exceed the performance of fully supervised models on near-distribution outliers[91,92].

In summary, GedankenNet overcomes important limitations of existing deep learning models in holographic microscopy by creating experimental-data-free, generalizable, and physics-compatible deep learning models. GedankenNet further opens up new opportunities for other microscopy imaging modalities and various computational inverse imaging problems and could facilitate a diverse set of applications for deep learning-based holography and microscopy techniques.

**Figures and figure legends**

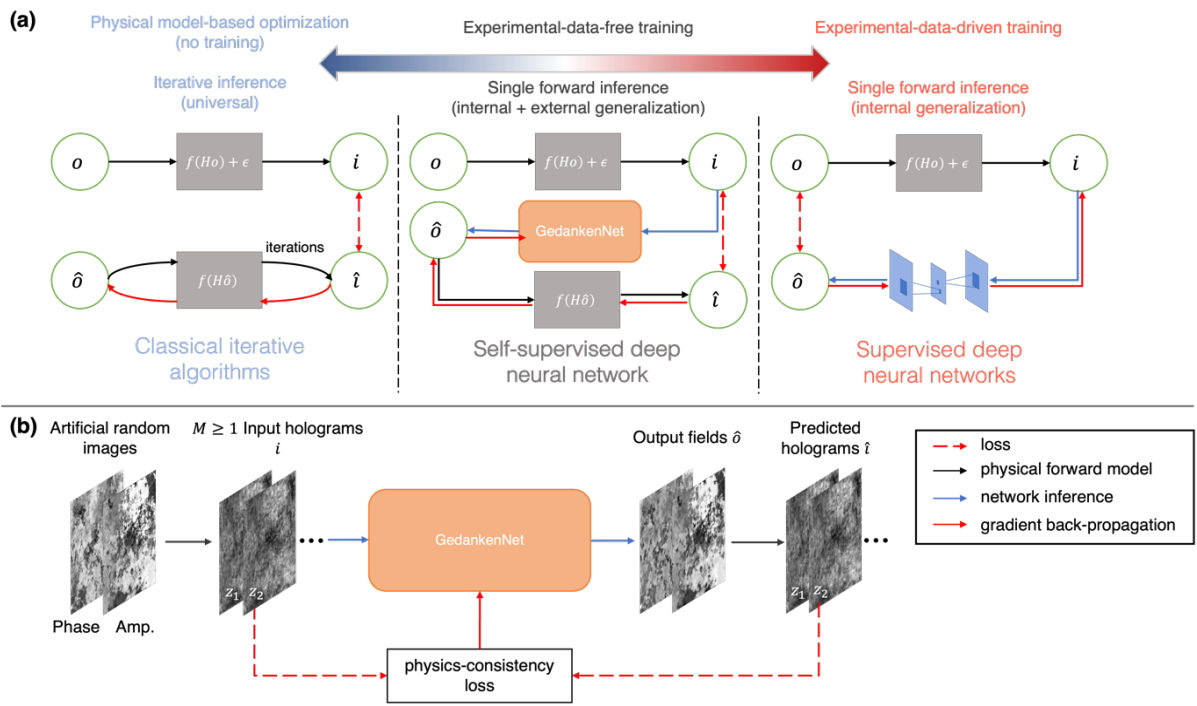

**Figure 1**. Diagrams of GedankenNet and other existing methods for solving holographic imaging problems. (a) Diagrams of classical iterative hologram reconstruction algorithms, the self-supervised deep neural network (GedankenNet) and existing supervised deep neural networks. (b) Self-supervised training pipeline of GedankenNet for hologram reconstruction.

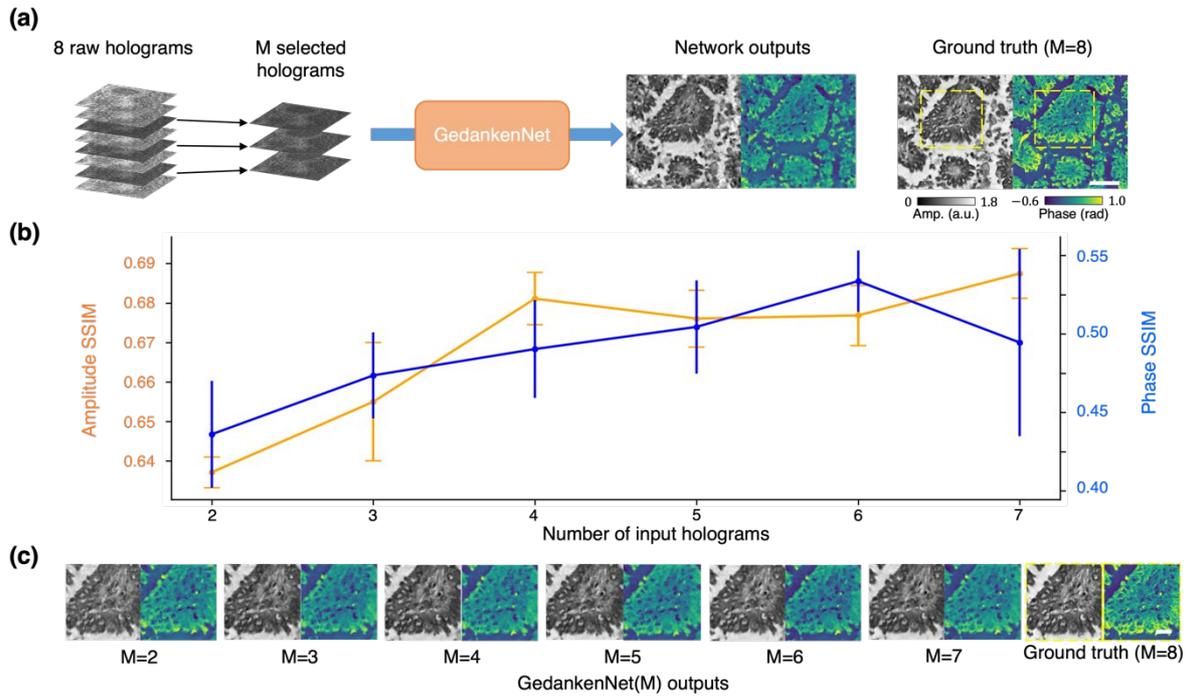

**Figure 2**. Hologram reconstruction performance of GedankenNet using multiple ($M$) input holograms. (a) $M$ holograms were selected from 8 raw holograms as the inputs for GedankenNet. The ground truth complex field (used only for comparison) was retrieved by MHPR using all the 8 raw holograms. Scale bar: 50 µm. (b) The amplitude and phase SSIM values between the reconstructed fields of GedankenNet and the ground truth object fields. SSIM values were averaged on a testing set with 94 unique human lung tissue FOVs, and the SSIM standard deviations were calculated on 4 individual models for each $M$. (c) Zoomed-in regions of the GedankenNet outputs and the ground truth object fields. Scale bar: 20 µm.

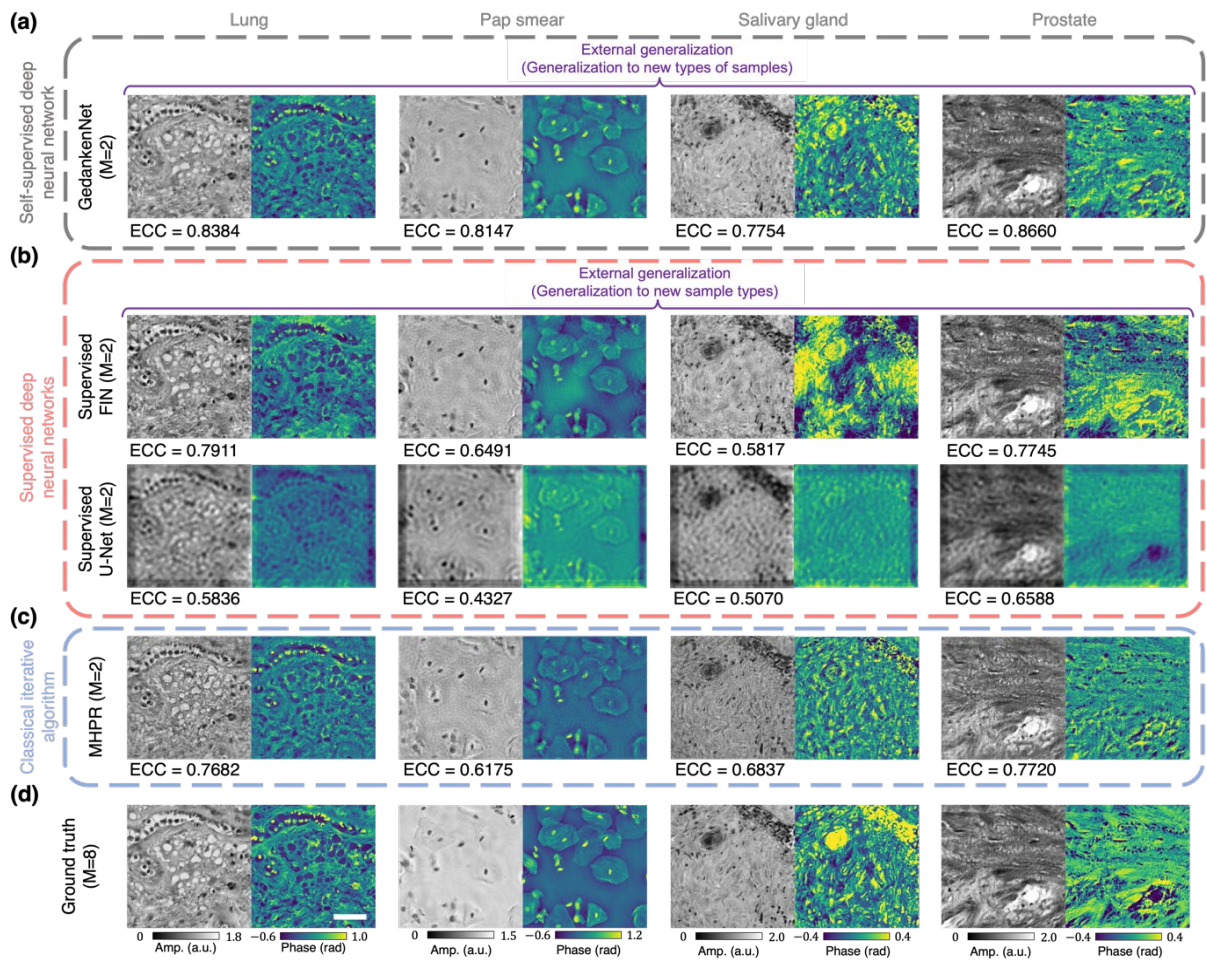

**Figure 3.** External generalization of GedankenNet on human tissue sections and Pap smears, and its comparison with existing supervised learning models and MHPR. (a) External generalization results of GedankenNet on human lung, salivary gland, prostate and Pap smear holograms. (b) External generalization results of supervised learning methods on the same test datasets. The supervised models were trained on the same simulated hologram dataset as GedankenNet used. (c) MHPR reconstruction results using the same *M* = 2 input holograms. (d) Ground truth object fields retrieved using 8 raw holograms of each FOV. Scale bar: 50 μm.

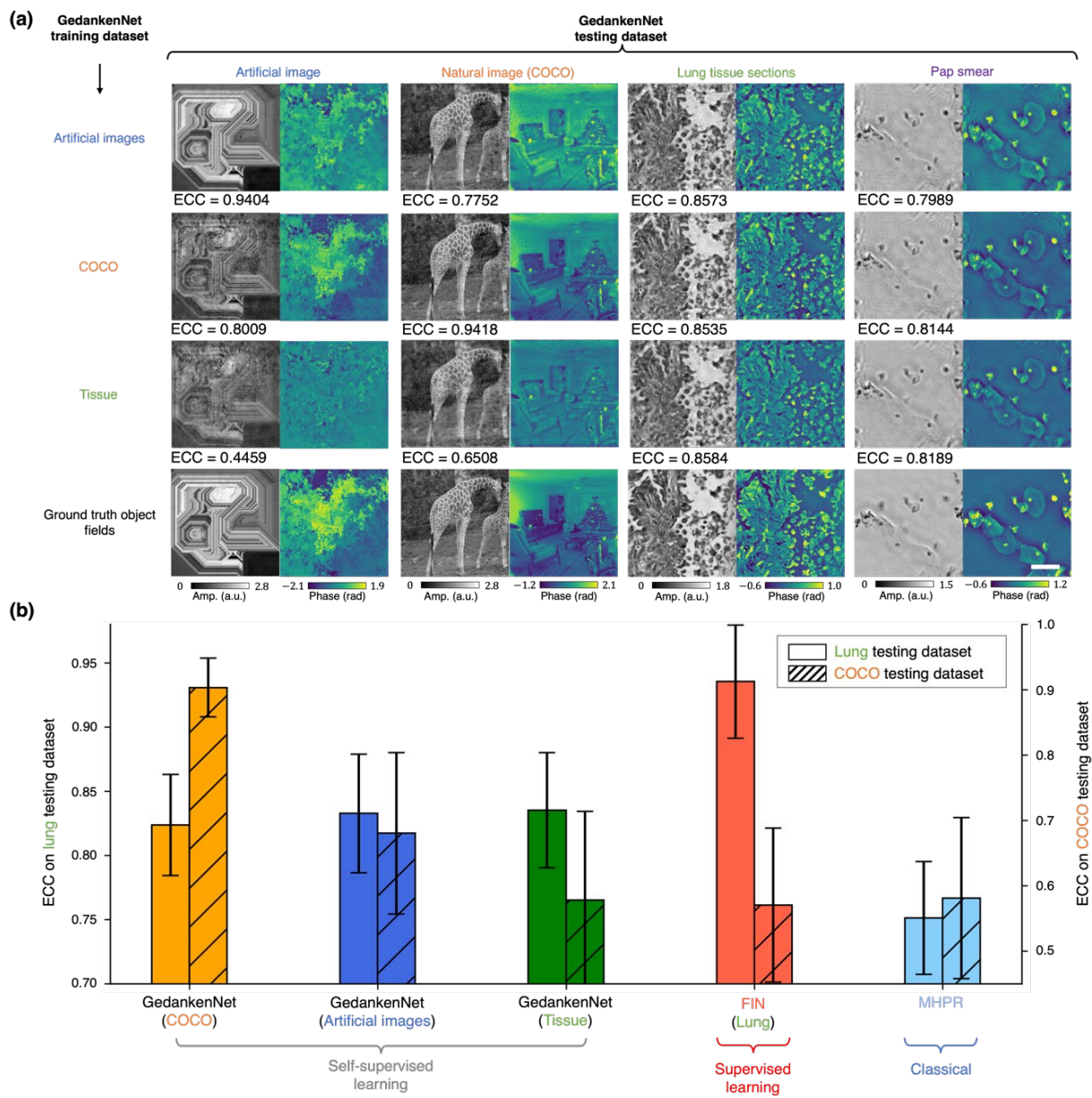

**Figure 4**. Generalization of GedankenNet trained with different training datasets to various new testing datasets. (a) Outputs of GedankenNet models trained on three different training datasets (artificially generated random synthetic images, natural images (COCO) and tissue sections, respectively). (b) Quantitative performance analysis of GedankenNet models trained on three different datasets. The performances of a supervised deep neural network (trained on lung tissue sections) and MHPR are also included for comparison purposes. ECC mean and standard deviation values were calculated on lung and COCO test datasets with 94 and 100 unique FOVs, respectively. Scale bar: 50 μm.

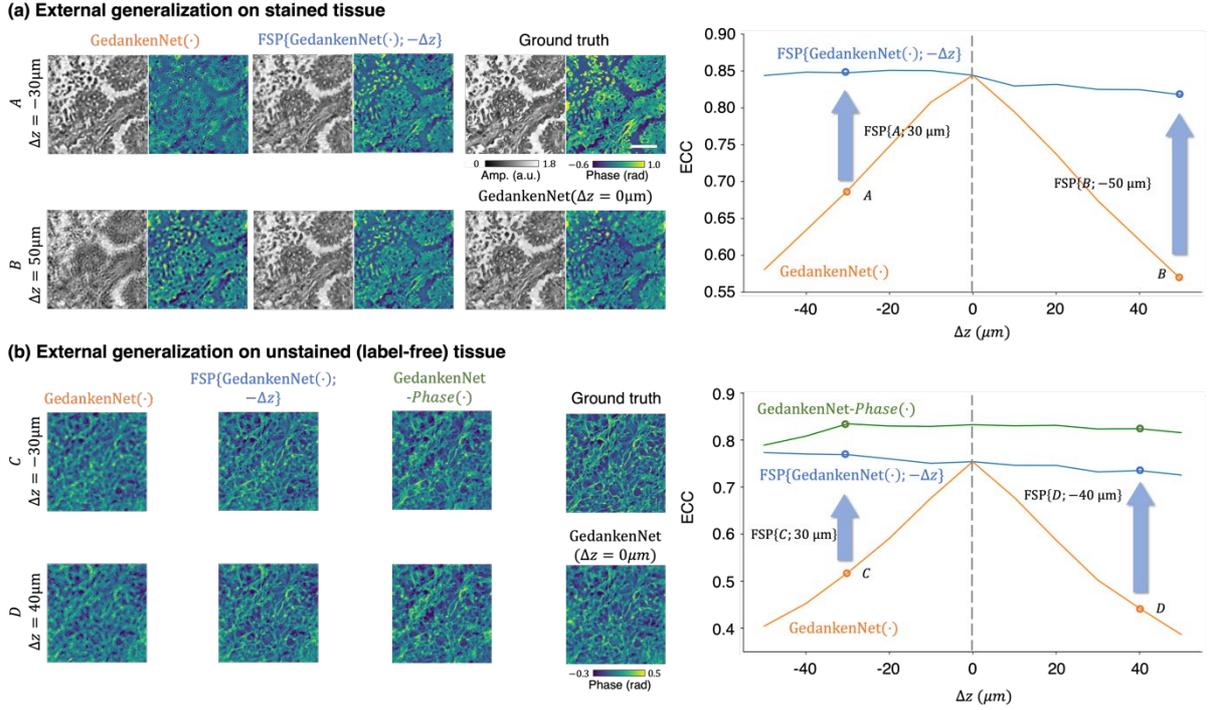

**Figure 5**. Compatibility of GedankenNet output images with the wave equation in free-space. GedankenNet model was trained to reconstruct $M = 2$ input holograms at $z_1 = 300$μm and $z_2 = 375$μm, but blindly tested on input holograms captured at $z'_1 = 300 + \Delta z$ μm, $z'_2 = 375 + \Delta z$ μm (orange curve). The resulting GedankenNet output complex fields are propagated in free-space by $-\Delta z$ using the wave equation, revealing a very good image quality (blue curve) across a wide range of axial defocus distances. The GedankenNet-*Phase* (green curve) was trained to reconstruct sample fields with $M = 2$ input holograms at arbitrary, unknown axial positions within $[275, 400]$ μm. (a) External generalization on stained human lung tissue sections. (b) External generalization on unstained, label-free human kidney tissue sections. These results demonstrate that GedankenNet framework not only has a superior external generalization to experimental holograms (using experiment- and data-free training), but also very well generalized to work with defocused experimental holograms, and encoded the wave equation into its inference process using the physics-consistency loss. Scale bar: 50 μm.

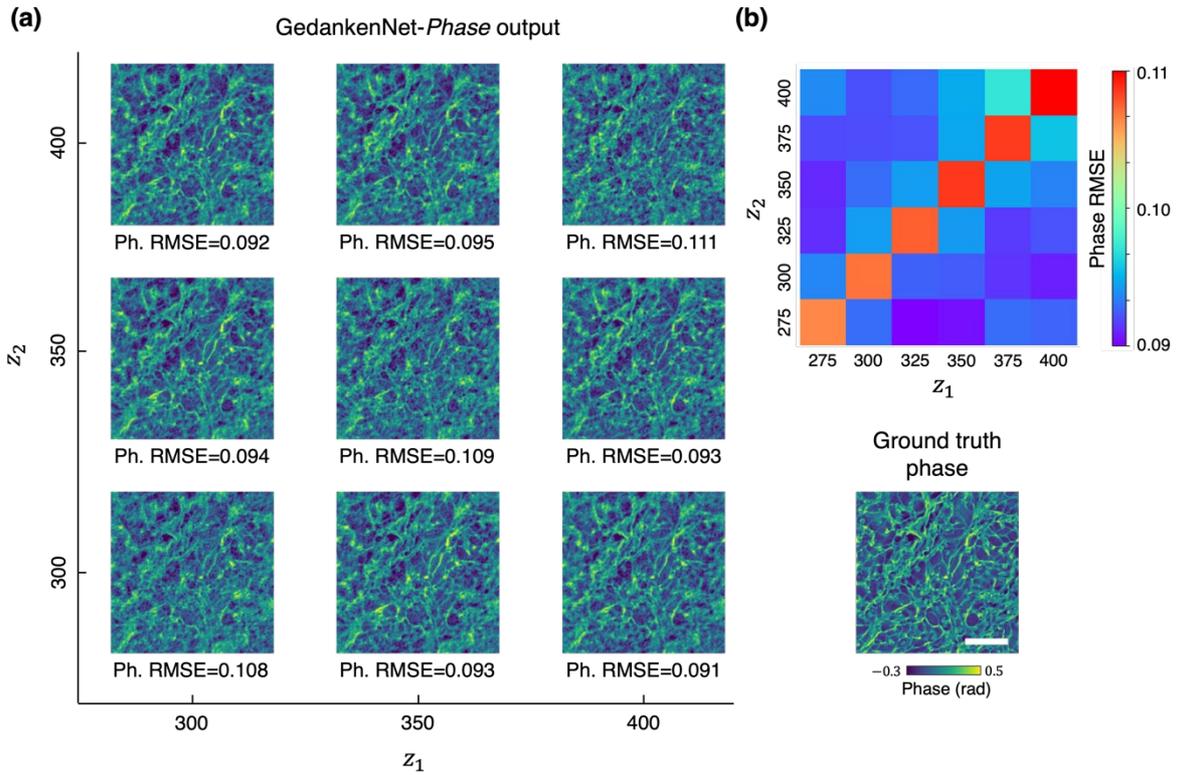

**Figure 6.** Autofocusing performance of GedankenNet-*Phase* for experimental holograms of unstained human kidney tissue sections captured at various sample-to-sensor distances. (a) Reconstructed sample phase by GedankenNet-*Phase*. (b) Phase RMSE of the reconstructed phase images by GedankenNet-*Phase* with respect to the ground truth. Phase RMSE values were averaged on the unstained kidney test dataset with 98 unique FOVs. The ground truth object fields were retrieved from 8 raw holograms. Scale bar: 50 μm.

**Methods**

**Sample preparation and imaging**

Human tissue samples used in this work were prepared and provided by the UCLA Translational Pathology Core Laboratory (TPCL). A fraction of tissue slides were stained with Hematoxylin and Eosin (H&E) to reveal structural features in the amplitude channel and the other slides remained unstained to serve as phase-only objects. Stained Pap smears were acquired from the UCLA Department of Pathology. All slides were deidentified and prepared from existing specimens without links or identifiers to the patients.

The experimental holograms were captured on stained human lung, prostate, salivary gland, kidney, liver and esophagus tissue sections, Pap smears, and unstained label-free human kidney tissue sections. Holographic microscopy imaging was implemented using a lens-free, in-line holographic microscope as illustrated in Extended Data Fig. 1(b). The custom-designed microscope was equipped with a tunable light source (WhiteLase Micro, NKT Photonics) and an acousto-optic tunable filter (AOTF). In the reported experiments, the AOTF was set to filter the illumination light at $\lambda_0 = 530$ nm wavelength unless otherwise specified. Raw holograms ($\sim 4.6\text{K} \times 3.5\text{K}$ pixels) were recorded by a complementary metal-oxide semiconductor (CMOS) with a pixel size of 1.12 μm (IMX 081 RGB, Sony Corp.). A 6-axis 3D positioning stage (MAX606, Thorlabs, Inc.) controlled the CMOS sensor to capture raw holograms consecutively for each FOV at various sample-to-sensor distances, which ranged from ~300 μm to ~600 μm in this work with an axial spacing of ~10-15 μm. A computer connected all the devices and automatically controlled the image acquisition process through a LabView script.

**Artificial hologram preparation and preprocessing**

The artificial holograms used in this work for the training were simulated either from random images or natural images (from COCO dataset). Random images (with no connection or resemblance to real-world samples) were generated using a Python package *randimage*, which colored the pixels along a path found from a random gray-valued image to generate an artificial RGB image. Then we mapped the generated random RGB images to grayscale. Two independent images randomly selected from the dataset served as the amplitude and phase of the complex object field, and a small constant was added to the amplitude channel to avoid zero transmission and undefined phase issue. For the artificial random phase-only object fields, only the phase image was selected, and the amplitude was set as 1 everywhere. The object field was then propagated by the given sample-to-sensor distances using the angular spectrum approach[83], and the intensity of the resulting complex field was calculated. The resulting holograms were cropped into $512 \times 512$ patches. Each of the two datasets (either from random images or COCO natural images) used ~100K images for training and a set of 100 images for validation and testing, which were excluded from training. All models in this work used the amplitude of the measured fields as the inputs.

Given a randomly selected amplitude ($A$) image and phase ($\phi$) image, the simulated hologram $i(x, y; z)$ at axial position $z$ is generated by free-space propagation (FSP):

$$i(x,y;z) = \left| \text{FSP}\left((A + \delta) \odot e^{i\pi\phi}; z\right) + \epsilon \right|$$

where $\delta \in \mathbb{R}$ stands for the added small constant, $\odot$ represents element-wise multiplication, and $\epsilon \in \mathbb{C}^{N \times N}$ is the additional white Gaussian noise. For the phase-only objects, the simulated holograms can be expressed as:

$$i(x,y;z) = |\text{FSP}(e^{i\pi\phi}; z) + \epsilon|$$

The FSP is implemented based on the angular spectrum propagation method[83] by taking into account all the traveling waves in free-space. The angular spectrum of a light field $U(x, y; z_0)$ at the axial position $z_0$ can be expressed as

$$\tilde{A}(\xi, \eta; z_0) = \mathcal{F}\{U(x, y; z_0)\}$$

The angular spectrum of the propagated field at $z$ is related to $A(\xi, \eta; z_0)$ by

$$\tilde{A}(\xi, \eta; z) = \begin{cases} e^{i(z-z_0)\sqrt{k^2-\xi^2-\eta^2}} \cdot \tilde{A}(\xi, \eta; z_0), & \text{if } \xi^2 + \eta^2 < k^2 \\ 0, & \text{otherwise} \end{cases}$$

Here $\mathcal{F}$ and $\mathcal{F}^{-1}$ are the fast Fourier transform (FFT) pairs (forward vs. inverse). $x, y$ and $\xi, \eta$ are spatial and frequency domain coordinates, respectively. $k$ is the wave number of the illumination light in the medium. The FSP then infers the propagated field at $z$ by:

$$\text{FSP}(U(x, y; z_0); z - z_0) = \mathcal{F}^{-1}\{\tilde{A}(\xi, \eta; z)\}$$

**Experimental hologram dataset preparation and processing**

Raw experimental holograms were pre-processed through pixel super-resolution and autofocusing algorithms to retrieve sub-pixel features of the samples. For this, a pixel super-resolution (PSR) algorithm[84,93] was applied to raw experimental holograms to obtain high-resolution holograms, resulting in a final effective pixel size of 0.37 μm. Then, an edge sparsity-based autofocusing algorithm[94] was employed to determine the sample-to-sensor distances for each super-resolved hologram. The ground truth sample field was retrieved from $M = 8$ super-resolved holograms of the same FOV using the MHPR algorithm[84–86]. The MHPR algorithm retrieves the sample complex field through iterations between 8 input holograms. The initial guess of the sample complex field is propagated to each measurement plane using FSP and the corresponding sample-to-sensor distance. Then, the propagated field is updated by replacing the amplitude with the measured one and retaining the phase. One iteration is completed after all 8 holograms have been used. The algorithm generally converges after 100 iterations.

Input-target pairs of $512 \times 512$ pixels were cropped from the super-resolved holograms and their corresponding retrieved ground truth fields, forming the experimental hologram datasets. Standard data augmentation techniques were applied, including random rotations by $0, \pm 45, \pm 90$ degrees and random vertical and horizontal flipping. The multi-height experimental hologram dataset of tissue sections contains ~100K input-target pairs of stained human lung, prostate, salivary gland, kidney, liver and esophagus tissue sections. A subset of the lung, prostate, salivary gland slides from new patients and Pap smears were excluded from the training dataset and used as testing datasets, containing 94, 49, 49 and 47 unique FOVs, respectively. The holograms of the unstained (label-free) kidney tissue thin sections (~3-4 μm thick) were used as our phase-only object test dataset containing 98 unique FOVs.

**Network architecture**

A sequence of *M* holograms is concatenated as the input image with *M* channels and the real and imaginary parts of the object complex field are generated at the output of GedankenNet. GedankenNet contains a series of spatial-Fourier transformation (SPAF) blocks and a large-scale residual connection, in addition to two $1 \times 1$ convolution layers at the head and the tail of the network (see Extended Data Fig. 1(a)). In each SPAF block, input tensors pass through two recursive SPAF modules with residual connections, which share the same parameters before entering the PReLU (parametric rectified linear unit) activation layer[95]. The PReLU activation function with respect to an input value $x \in \mathbb{R}$ is defined as:

$$\text{PReLU}(x) = \max(0, x) + a * \min(0, x)$$

where $a \in \mathbb{R}$ is a learnable parameter. Another residual connection passes the input tensor after the PReLU layer. The SPAF module consists of a 3 × 3 convolution layer and a branch performing linear transformation in the Fourier domain (Extended Data Fig. 1(a)). The input tensor with $c$ channels to the SPAF module is first transformed into the frequency domain by a 2D FFT and truncated by a window with a half size $k$ to filter out higher frequency components. The linear transformation in the frequency domain is realized through pixel-wise multiplication with a trainable weight tensor $W \in \mathbb{R}^{c \times (2k+1) \times (2k+1)}$, i.e.,

$$F'_{j,u,v} = W_{j,u,v} \cdot \sum_{i=1}^{c} F_{i,u,v}, \quad u, v = 0, \pm 1, \ldots, \pm k, \quad j = 1, \ldots, c$$

where $F \in \mathbb{C}^{c \times (2k+1) \times (2k+1)}$ are the truncated frequency components. The resulting tensor $F'$ is then transformed into the spatial domain through an inverse 2D FFT. The same pyramid-like setting of half window size $k$ as in Ref. [81] was applied here such that $k$ decreases for deeper SPAF blocks. This pyramid-like setting provides a mapping of the high-frequency information of the holographic diffraction patterns to low-frequency regions in the first few layers and passes this low-frequency information to the subsequent layers with a smaller window size, which better utilizes the spatial features at multiple scales and at the same time considerably reduces the model size, avoiding potential overfitting and generalization issues.

The architecture of GedankenNet was extended for two additional models reported in the Results section, namely GedankenNet-*Phase* and GedankenNet-*Phaseλ*, as shown in the Extended Data Figure 8(a). Similar to GedankenNet, these models use a sequence of $M$ holograms concatenated as the input image with $M$ channels, but, instead of outputting real and imaginary parts, the GedankenNet-*Phase* and GedankenNet-*Phaseλ* only generate phase-only output images. The Dynamic SPAF (dSPAF) modules[96] inside GedankenNet-*Phase* and GedankenNet-*Phaseλ* exploit a shallow U-Net to dynamically generate weights $W$ for each input tensor, and enable the capabilities of autofocusing and adapting to unknown shifts/changes in the illumination wavelengths. The dense links provide an efficient flow of information from the input layer to the output layer, so that every output tensor of the dSPAF group is appended and fed to the subsequent dSPAF groups, resulting in an economic and powerful network architecture.

**Algorithm implementation**

GedankenNet, GedankenNet-*Phase* and GedankenNet-*Phase*$\lambda$ were implemented using PyTorch[97]. We calculated the loss values based on the hologram amplitudes, i.e.:

$$\hat{\imath} = |\text{FSP}(\hat{o}; z_1, z_2, \cdots, z_M)|$$

The training loss consists of three individual terms: (1) FDMAE loss between the predicted holograms $\hat{\imath}$ and the input holograms $i$; (2) MSE loss between $\hat{\imath}$ and $i$; and (3) TV loss on the output complex field $\hat{o}$. The first two terms constitute the physics-consistency loss, and the total loss is a linear combination of the three terms, expressed as:

$$L_{total} = L_{physics-consistency}(\hat{\imath}, i) + \gamma L_{TV}(\hat{o})$$
$$= \alpha L_{FDMAE}(\hat{\imath}, i) + \beta L_{MSE}(\hat{\imath}, i) + \gamma L_{TV}(\hat{o})$$

where $\alpha, \beta, \gamma$ are loss weights empirically set as 0.1, 1, and 20.

The FDMAE loss is calculated as:

$$L_{FDMAE}(\hat{\imath}, i) = \frac{1}{N^2} \sum_{\xi=1}^{N} \sum_{\eta=1}^{N} |\mathcal{F}\{\hat{\imath}\}(\xi, \eta) \cdot w(\xi, \eta) - \mathcal{F}\{i\}(\xi, \eta) \cdot w(\xi, \eta)|$$

Here $w \in \mathbb{R}^{N \times N}$ is a 2D Hann window[98], and $\xi, \eta$ are indices of frequency components. MSE and TV losses are computed using:

$$L_{MSE}(\hat{\imath}, i) = \frac{1}{N^2} \sum_{x=1}^{N} \sum_{y=1}^{N} |\hat{\imath}(x, y) - i(x, y)|^2$$

$$L_{TV}(\hat{o}) = \frac{1}{2N^2} \sum_{x=1}^{N} \sum_{y=1}^{N} |\nabla_x Re\{\hat{o}\}(x,y)| + |\nabla_y Re\{\hat{o}\}(x,y)| + |\nabla_x Im\{\hat{o}\}(x,y)|$$
$$+ |\nabla_y Im\{\hat{o}\}(x,y)|$$

Here $x, y$ are spatial indices, $\nabla_x, \nabla_y$ refer to the differentiation operation along the horizontal and vertical axes, $Re\{\cdot\}, Im\{\cdot\}$ return the real and imaginary parts of the complex fields, respectively.

For the GedankenNet-*Phase* and GedankenNet-*Phaseλ*, which only generate phase-only output fields, the predicted hologram was calculated using:

$$\hat{\imath} = |FSP(e^{i\pi\hat{p}}; z_1, z_2, \cdots, z_M)|$$

where $\hat{p}$ is the output phase field. The TV loss was calculated by using:

$$L_{TV}(\hat{p}) = \frac{1}{N^2} \sum_{x=1}^{N} \sum_{y=1}^{N} |\nabla_x \hat{p}(x,y)| + |\nabla_y \hat{p}(x,y)|$$

To avoid trivial ambiguities in phase retrieval[99–101], the GedankenNet's output was normalized using its complex mean; the GedankenNet-*Phase* and GedankenNet-*Phaseλ*'s outputs were subtracted from their corresponding mean.

All the trainable parameters in GedankenNet were optimized using the Adam optimizer[102]. The learning rate follows a cosine annealing scheduler with an initial rate of 0.002. All the models went through ~ 0.75 million batches (equivalent to ~7.5 epochs) and the best model was preserved with the minimal validation loss. The training takes ~48 hours for an $M = 2$ model on a computer equipped with an i9-12900F CPU, 64 GB RAM and an RTX 3090 graphics card. The inference time measurement (Extended Data Table 1) was done on the same machine with GPU acceleration and a test batch size of 20 for GedankenNet, 12 for both GedankenNet-*Phase* and GedankenNet-*Phaseλ*.

The supervised FIN adopted the same architecture and parameters as in Ref. [48]. The U-Net architecture employed four convolutional blocks in the down-sampling and up-sampling paths separately, and each block contained two convolutional layers with batch normalization and ReLU activation. The input feature maps of the first convolutional block had 64 channels and each block in the down-sampling path doubled the number of channels. Supervised FIN and U-Net[87] models adopted the same loss function as in Ref. [48]. The same Adam optimizer and learning rate were applied to the supervised learning models. DIP adopted a U-Net architecture, an Adam optimizer and the loss function used in Ref. [80].

**Image reconstruction evaluation metrics**

SSIM, RMSE and ECC were used in our work to evaluate the reconstruction quality of the output fields with respect to the ground truth fields. SSIM and RMSE are based on single-channel images. Denote $\hat{o} \in \mathbb{R}^{N \times N}$ as the reconstructed amplitude or phase image, and $o \in \mathbb{R}^{N \times N}$ as the ground truth amplitude or phase image, SSIM and RMSE values were calculated using the following equations:

$$\text{SSIM}(\hat{o}, o) = \frac{(2\mu_{\hat{o}}\mu_o + c_1)(2\sigma_{\hat{o}o} + c_2)}{(\mu_{\hat{o}}^2 + \mu_o^2 + c_1)(\sigma_{\hat{o}}^2 + \sigma_o^2 + c_2)}$$

$$\text{RMSE}(\hat{o}, o) = \sqrt{\frac{1}{N^2} \sum_{x=1}^{N} \sum_{y=1}^{N} (\hat{o}(x,y) - o(x,y))^2}$$

Here $\mu_{\hat{o}}, \mu_o$ stand for the mean of $\hat{o}, o$, respectively. $\sigma_{\hat{o}}^2, \sigma_o^2$ stand for the variance of $\hat{o}, o$, respectively, and $\sigma_{\hat{o}o}$ is the covariance between $\hat{o}$ and $o$. $c_1 = 2.55^2, c_2 = 7.65^2$ are constants used for 8-bit images. $x, y$ are 2D coordinates of the image pixels.

The ECC is calculated based on the reconstructed complex field and the ground truth field. $\hat{o}' \in \mathbb{C}^{N \times N}$ is the reconstructed field obtained by subtracting $\hat{o}$ with its mean value. $o' \in \mathbb{C}^{N \times N}$ is the corresponding ground truth field. The ECC can be calculated as:

$$ECC(\hat{o}', o') = Re\left\{\frac{vec(\hat{o}')^H \cdot vec(o')}{\|vec(\hat{o}')\| \cdot \|vec(o')\|}\right\}$$

Here $vec(\hat{o}')^H$ is the conjugate transpose of the vectorized $\hat{o}'$, and $\|\cdot\|$ is the Euclidean norm.